\documentclass[10pt,twocolumn,letterpaper]{article}

\usepackage[pagenumbers]{cvpr} 

\usepackage{graphicx}
\usepackage{amsmath}
\usepackage{amssymb}
\usepackage{booktabs}
\usepackage{mathtools}
\usepackage{amsthm}
\usepackage{multirow}
\usepackage{subcaption}
\usepackage{bm}
\usepackage{xspace}
\usepackage{enumitem}
\usepackage{wrapfig}

\usepackage[pagebackref,breaklinks,colorlinks]{hyperref}

\usepackage[capitalize]{cleveref}
\crefname{section}{Sec.}{Secs.}
\Crefname{section}{Section}{Sections}
\Crefname{table}{Table}{Tables}
\crefname{table}{Tab.}{Tabs.}

\theoremstyle{plain}

\theoremstyle{definition}

\theoremstyle{remark}

\usepackage[textsize=tiny]{todonotes}

\usepackage{xspace}
\usepackage[normalem]{ulem}
\useunder{\uline}{\ul}{}
\usepackage{booktabs}

\def\Eq#1{Eq.~(\ref{eq:#1})}

\def\be{\begin{equation}}
\def\ee{\end{equation}}
\def\bea{\begin{eqnarray}}
\def\eea{\end{eqnarray}}
\def\fig#1{Figure~\ref{fig:#1}}
\def\tab#1{Table~\ref{tab:#1}}
\def\sect#1{Section~\ref{sec:#1}}

\def\R{{\rm I\!R}}

\makeatletter
\DeclareRobustCommand\onedot{\futurelet\@let@token\@onedot}
\def\@onedot{\ifx\@let@token.\else.\null\fi\xspace}
\def\eg{{e.g}\onedot, }

\def\ie{{i.e}\onedot, }

\def\vs{{vs}\onedot} 

\def\wrt{w.r.t\onedot} 
\def\etal{\emph{et al}\onedot}

\def\mypar#1{\vspace{1mm}{\noindent\bf #1}\hspace{1mm}}

\newcommand{\myskip}[1]{} 
\newcommand{\myres}[1]{{$#1 \times #1$}}

\begin{document}

\title{Multi-Domain Learning with Modulation Adapters}

\author{Ekaterina Iakovleva\\
Inria\\
\and
Karteek Alahari\\
Inria\\
\and
Jakob Verbeek\\
Meta AI\\
}
\maketitle

\begin{abstract}
Deep convolutional networks are ubiquitous in computer vision, due to their excellent performance across different tasks for various domains. 
Models are, however, often trained in isolation for each task, 
failing to exploit relatedness between tasks and domains to learn more compact models that generalise better in low-data regimes.
Multi-domain learning aims to handle related tasks, such as image classification across multiple domains, simultaneously.
Previous work on this problem explored the use of a pre-trained and fixed domain-agnostic base network, in combination with smaller learnable domain-specific adaptation modules. 
In this paper, we introduce Modulation Adapters, which update the convolutional filter weights of the model in a multiplicative manner for each task.
Parameterising these adaptation weights in a factored manner  allows us to scale the number of per-task parameters in a flexible manner, and to strike different parameter-accuracy trade-offs.
We evaluate our approach on the Visual Decathlon challenge, composed of ten image classification tasks across different domains, and on the ImageNet-to-Sketch benchmark, which consists of six image classification tasks.
Our approach yields excellent results, with accuracies that are comparable to or better than those of existing state-of-the-art approaches. 
\end{abstract}

\section{Introduction}
\label{sec:introduction}

Deep learning models are ubiquitous across many tasks and application domains~\cite{lecun15nature}. 
Yet, one of their biggest limitations is the reliance on large amounts of labelled data  to train these models, which often have millions if not billions of parameters.
Knowledge transfer is one of the main solutions to this problem, where knowledge obtained by learning to solve one task is leveraged to learn another task in a more sample-efficient manner.
A popular example of knowledge transfer is pre-training of a deep neural network on a large (un)labelled dataset, \eg on ImageNet~\cite{russakovsky15ijcv}, and then fine-tuning it on a smaller target dataset. 

Beyond pre-training and fine-tuning, there is a wide range of learning paradigms that involve some form of knowledge transfer. 
In \emph{multi-domain learning} (MDL)~\cite{rebuffi17nips}, which is the focus of this paper, a single model is trained to solve several related tasks on a number of different domains, \eg digit recognition and object recognition. The idea is to use knowledge from related tasks to induce regularisation across the considered domains. 
The common view on MDL is to transfer knowledge across domains to maximise predictive accuracy, while  sharing as many parameters as possible across tasks, and adding as few   task-specific parameters as possible. 
A common approach is to learn a base network on one of the domains, and then adapt it with a small number of parameters for tasks on other domains.
A wide range of solutions to such base network adaptations has been proposed in the literature, which can be divided into two major groups of approaches. 
The first group applies a task-specific binary mask to either the features  or the weights in each layer of the model~\cite{berriel19iccv,mallya18eccv,mancini18eccv}.
While this  adds a small number of parameters per task, essentially one bit per weight or feature channel, its ability to adapt to new tasks is limited for the same reason.
The second group leaves the weights of the  base network unchanged, but adds a number of task-specific adaptation layers, for example, a \myres{1} convolution  parallel to each \myres{3} convolutational layer of the base network~\cite{li19cvpr,rebuffi17nips,rosenfeld18pami}.  
Depending on the design of the adaptation modules, this approach adds more parameters to adapt the model to the task. However, by leaving the parameters of the base network layers unchanged, task adaptation can only be achieved by added model capacity in the adaptation layers themselves, without capitalising on adaptations of all of the backbone parameters. 

\begin{figure*}[ht]
\centering
\hspace{30pt}
\begin{subfigure}{.25\textwidth}
    \centering
    \includegraphics[width=.6\textwidth]{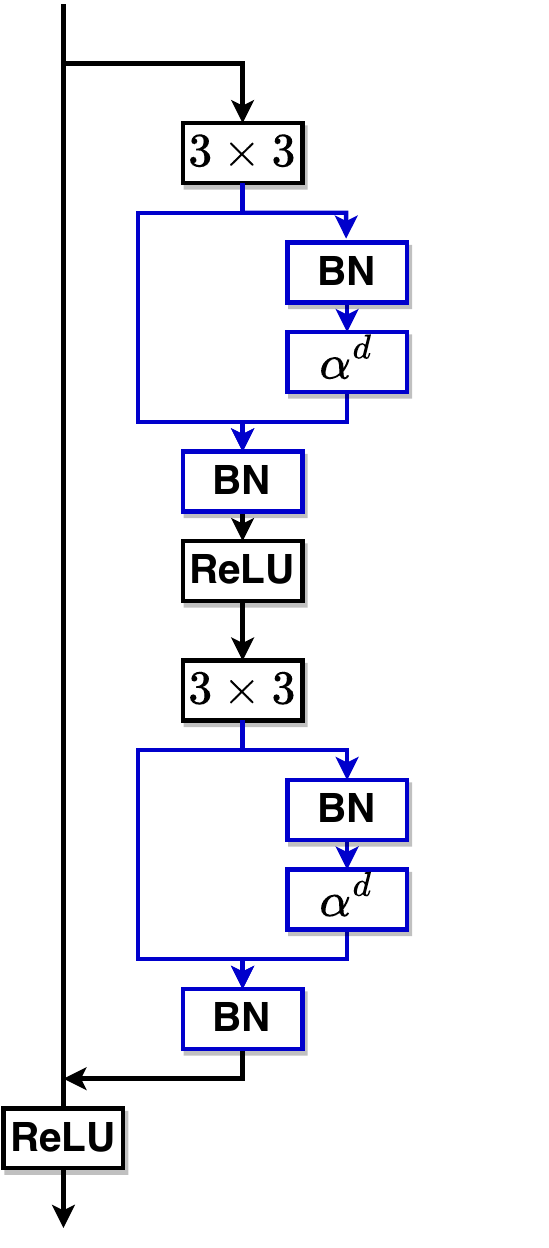}
    \hspace{-70pt}\caption{Residual Adapters}
    \label{fig:residual_adapters}
\end{subfigure}
\hfill
\begin{subfigure}{.25\textwidth}
    \centering
    \includegraphics[width=.6\textwidth]{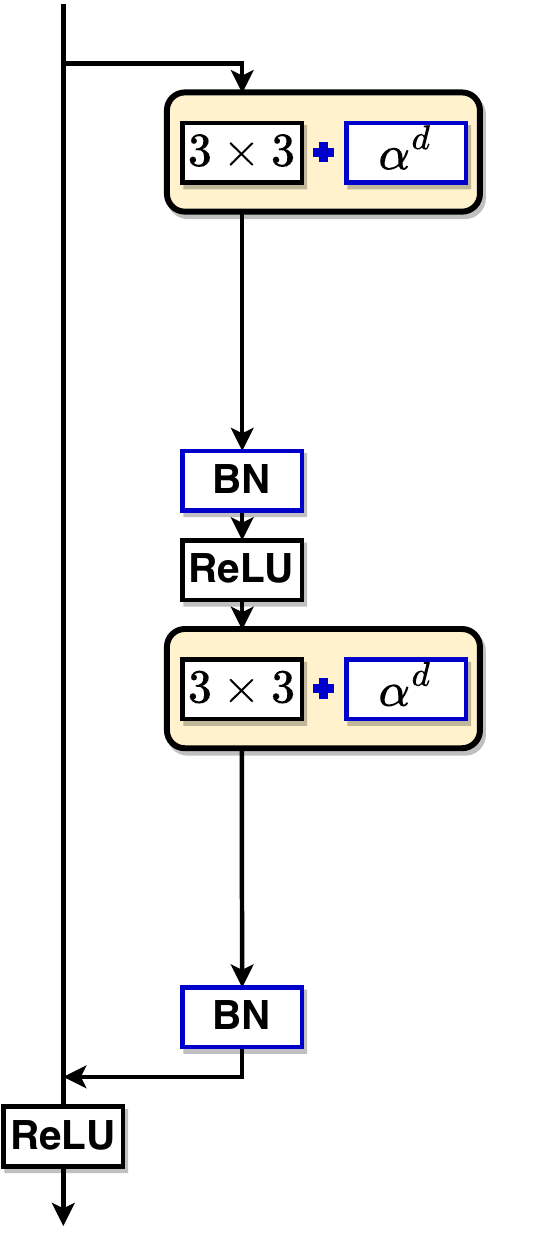}
    \caption{Parallel Adapters}
    \label{fig:parallel_adapters}
\end{subfigure}
\hfill
\begin{subfigure}{.25\textwidth}
    \centering
    \includegraphics[width=.6\textwidth]{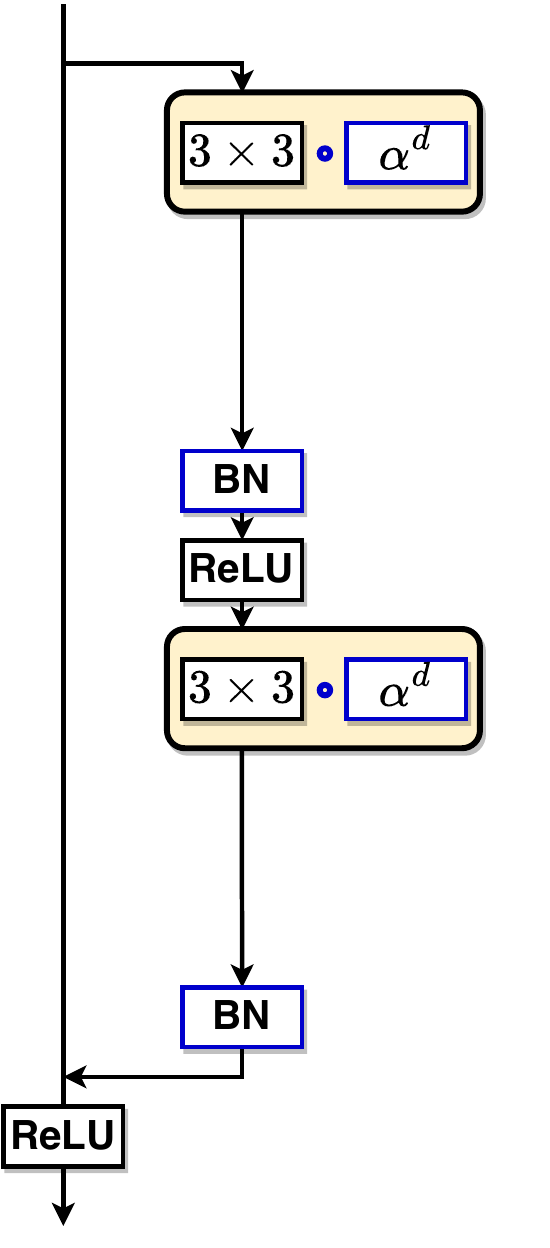}
    \caption{Modulation Adapters}
    \label{fig:modulation_adapters}
\end{subfigure}
\hspace{30pt}
\caption{Different adaptation units embedded into a residual block. 
Blocks in blue, adapters ${\bm\alpha}^{d}$ and batch-norm (BN) layers,  contain domain-specific parameters. 
Blocks in black in the base network are domain-agnostic. Yellow blocks show that the $3\times3$ convolutions of the base network are modulated before being applied. 
Note that the BN blocks contribute a negligible number of parameters compared to the Modulation Adapter blocks.
}
\label{fig:adapted_residual_block}
\end{figure*}

To overcome the limitations of  existing approaches, we propose a novel type of adaptation module, which we call \emph{Modulation Adapters}. 
Our approach adapts the weights of the existing layers of the base network to the task by means of non-binary scaling.
More specifically, to adapt the base network to a new domain $d$, we learn a set of domain-specific adapters ${\bm\alpha}^d$, each modulating the fixed convolutional filters by multiplying them with a scalar $\alpha^d_{mn}$ that is specific to the pair $(m,n)$ of the output and input channels to which the filter applies. 
See \fig{modulation_adapters} for an illustration.
The latter allows for nine-fold parameter reduction compared to  \myres{3} convolutions that constitute the core of common  base networks, such as 
ResNets~\cite{he16eccv}.
To further reduce the memory footprint, we propose a factorised representation of each adapter ${\bm \alpha}^d$ as a  product of two matrices with a smaller intermediate dimensionality. 
In contrast to other approaches, our  adapters allow to modulate all weights in the base network in a non-binary manner, while offering a scalable number of parameters. 

We perform several ablative studies to assess  the impact of multiplicative adaptation compared to additive, and demonstrate the benefits of the proposed approach. We evaluate our Modulation Adapters on the Visual Decathlon Challenge and the ImageNet-to-Sketch benchmark.
We obtain excellent results, with accuracies that are comparable or better than those of existing state-of-the-art approaches across a wide range of parameter budgets. 

\section{Related work}
\label{sec:related_work}

\mypar{Weight and feature masking.}
\label{sec:masking}
One way to approach multi-domain learning is to adapt the convolutional layers  of a base network that has been pre-trained on a large dataset. The ``piggy-back'' approach \cite{mallya18eccv} learns element-wise binary masks for the weights in the convolutional layers of the base network. Such parameterisation allows memory-efficient storage of domain-specific parameters: requiring to store 1 bit for each real-valued weight, \ie a 32 fold reduction in storage for single-precision parameters.
The idea of adaptation through binary masks is further generalised in WTPB~\cite{mancini18eccv}, where the authors use three scalars to define an affine transformation of convolutional filter weights based on the binary masks. 
In BA$^2$~\cite{berriel19iccv}, binary masks are applied across feature channels rather than network weights.
As a result, after training the masked feature maps can be removed from the computational graph, reducing both the memory footprint and the computational complexity.
While reduction in memory budget is an appealing property of the masking approach, the common downside is limited capability of model adaptation due to simplicity of the binary masks. 
In our work, we propose a method for non-binary multiplicative adaptation of the base network, which allows to span a large range of parameter budgets. 

\mypar{Adaptation modules.}
\label{sec:adapters}
Domain-specific trainable  modules to adapt a pre-trained base network have been explored in several methods.
Residual Adapters~\cite{rebuffi17nips} use an intermediate residual block with trainable \myres{1} convolution after each pre-trained and fixed \myres{3} convolution. Compared to the base network, such types of adapters lead to nine-fold reduction in the number of additional parameters learned for a new domain. In Parallel Adapters~\cite{rebuffi18cvpr}, 
a domain-specific \myres{1} convolution is used as  an additive bypass to each domain-agnostic \myres{3} convolution. 
The Parallel Adapter configuration provides more flexibility than the sequential one, since it can be used on top of existing pre-trained networks as the adapters need not be present when pre-training. 
DAN~\cite{rosenfeld18pami} learns a linear transformation of the output filters in the base convolutions, which is equivalent to insertion of intermediate \myres{1} convolution after each base convolution. Unlike Residual Adapters, this model does not use additional skip connections and batch normalisation layers during adaptation. CovNorm \cite{li19cvpr} learns approximation of adaptation layers which consists of whitening, mini-adaptation and colouring operations. 
For each trained adapter, the whitening and colouring matrices are obtained from PCA of corresponding input and output activations, while the mini-adaptation layer is learned from scratch. 
Our method also falls into the category of models which contain adaptation modules. 
Different from the other approaches, we perform modulation of fixed convolutional filters  by scaling their weights with a non-binary factor specific to each input-output channel combination.

Adapter modules have been investigated in the context of NLP~\cite{houlsby19icml} to specialize Transformer models~\cite{vaswani17nips} to specific tasks.
In the context of generative image models, adapters have been used to adapt pre-trained GANs to new domains with few samples~\cite{noguchi19iccv,robb20arxiv}. 
Our model is related to weight modulation approaches for GANs used in StyleGAN2~\cite{karras20cvpr} and in CISP~\cite{anokhin21cvpr}.

\mypar{Other approaches.}
\label{sec:others}
Besides the two main lines of work described above, several other directions have been explored.
SpotTune \cite{guo19iccv} uses a policy network that for each data sample from a new domain, and for each convolutional layer, predicts whether the weights should be fine-tuned or re-used from the pre-trained network. 
There is, however, no incentive to reduce the number of parameters.
In the work of Guo \etal \cite{guo19aaai}, the authors replace standard convolutions in the base network with depth-wise separable convolutions~\cite{chollet17cvpr}. Parameter-heavy \myres{1} convolutions are then shared across domains, while the per-feature \myres{3} convolutions with less parameters per task are learned. Such parameterisation of convolutions significantly decreases the size of the base network, as well as the number of parameters learned for each domain. 
Approach of Senhaji \etal \cite{senhaji21icpr} divides the fixed pre-trained base network with Parallel Adapters learned for a new domain into three non-overlapping blocks, and place an exit module after each of them. Each exit module takes the output of the previous block as input, and produces the softmax output for the domain of interest. The authors learn all exit modules and choose the one with the highest performance.

\section{Method}
\label{sec:method}

We consider the multi-domain learning problem where a model is required to solve $D$ classification tasks on $D$ data domains, while minimising the total number of parameters being learned. We begin with the popular framework, \eg \cite{mallya18eccv, mancini18eccv, rebuffi17nips}, of pre-training a base feature extractor on a large dataset, which is then fixed and shared across all domains. 
We then learn a separate adapter for each domain, which modifies the feature extractor to improve its performance on the given domain.
Along with the adapters, we also learn  domain-specific batch normalisation layers and a classification head which predicts  the  domain-specific labels. The objective function is a sum of domain-specific objective functions that correspond to individual tasks: in our case the  sum of corresponding cross-entropy losses.

We detail our approach and its properties in \sect{modulation_adapters}, and discuss the factored parametrisation in \sect{factorisation}. Then, in \sect{g_comparison}, we provide a detailed comparison of our approach with respect to the most related prior works.

\subsection{Modulation Adapters}
\label{sec:modulation_adapters}

We suppose the CNN base network has already been pre-trained. 
Let $\mathbf{f} \in \R^{M \times N \times K \times K}$ be the filter of a specific layer in the base network, where $K$ is the spatial filter size, $N$ is the number of input features, and $M$ is the number of output features. 

To adapt the convolutional layers of the base network to a new domain $d \in \{1,\ldots,D\}$, we define 
a modulation adaptation unit $\rho$, with domain specific parameters  ${\bm \alpha}^{d} \in \R^{M \times N}$ that modulate the convolutional filter in a multiplicative manner.
For an input $\mathbf{x}$ with $N$ feature maps, this unit produces an output $\mathbf{y}$ with $M$ feature maps:
\begin{equation}
\label{eq:modulation_adaptation_unit}
    \mathbf{y} = \rho(\mathbf{f} , \mathbf{x}, {\bm \alpha}^{d}) = 
    \left(\mathbf{f} \circ {\bm\alpha}^{d} \right) \ast \mathbf{x},
\end{equation}
where $\ast$ is the convolutional operator, and $\circ$ is an element-wise product with the adapter parameters ${\bm\alpha}^{d} \in \R^{M \times N}$ broadcasted across the spatial dimensions of the convolutional weight tensor $\mathbf{f} \in \R^{M \times N \times K \times K}$. 
Adaptation module $\rho$ applied to the pre-trained filter $\mathbf{f}$ is equivalent to a new convolutional filter $\mathbf{g}^{d}$, with the weights obtained as:
\begin{equation}
\label{eq:mad_new_convolution}
    [\mathbf{g}]^{d}_{mnkl} = \alpha^{d}_{mn} \cdot [\mathbf{f}]_{mnkl}.
\end{equation}
Note that modulation of the filter weights for each domain can also be interpreted as applying scaling of the input features, which is specific for each output channel. 
Therefore, with modulation each output channel can mix the input channels in different ways, \eg amplify some and ignore others. The input channels are, however, processed by the same spatial filter for each input-output channel combination, leading to reduction in number of trainable parameters relative to the size of the base network.

We learn a separate set of modulation weights for each convolutional layer in the base network, and denote our adaptation unit as a \emph{Modulation Adapter} (MAD). \fig{modulation_adapters} shows how Modulation Adapter units are incorporated into a single residual block. 
While all convolutional weights are altered for a new domain, we only need to store the pre-trained filters $\mathbf{f}$ as well as the Modulation Adapters, which are learned from scratch. 
The latter contain $MN$ parameters  
for each convolution in the base model. This is a reduction in the number of parameters by a factor $K^2$ (the spatial filter size), relative to corresponding convolution in the base network (e.g., for \myres{3} convolutions $K^2 = 3\times 3=9$).

\subsection{Factorisation of Modulation Adapters} 
\label{sec:factorisation}
The number of parameters in our Modulation Adapters is given by the product of the number of input channels and the number of output channels of a convolutional layer. 
For residual blocks the number of input and output channels is the same, and thus the number of adaptation parameters is quadratic in the number of channels.
Therefore, the number of adaptation parameters is typically largest in deeper layers of the base network which tend to contain hundreds of feature channels.
For example, the ResNet-26 architecture we use in our experiments has 256 channels in the deeper layers, resulting in 64k parameters in each adaptation module.
An alternative, allowing for linear scaling of the number of parameters with the number of feature channels, would be to simply scale the input and output feature channels of the convolutional layer. 
This scaling, however, could be absorbed in the domain specific BatchNorm layers, which has been found to be too restrictive for effective adaptation in previous work~\cite{rebuffi17nips}.

Inspired by parameter reduction demonstrated by matrix decomposition in CovNorm~\cite{li19cvpr}, we adopt a similar strategy in our work. However, instead of learning an unconstrained full adapter and approximating it with a product of two (or more) matrices using some form of matrix decomposition, e.g. SVD as in ConvNorm, we propose to directly represent and learn MAD as a product of two matrices, each with a smaller intermediate dimension $I < \min (M, N)$:

\begin{equation}
\label{eq:mad_decomposition}
    {\bm \alpha}^{d} = {\bm \beta}^{d} \times {\bm \gamma}^{d},
\end{equation}
where ${\bm\beta}^{d} \in \R^{M \times I}$, ${\bm\gamma}^{d} \in \R^{I \times N}$ and $\times$ denotes matrix multiplication. Such representation limits the maximum rank of the MAD to $I$, and can be interpreted as a rank factorisation. In practice, we observe that for certain base networks the full Modulation Adapters are, indeed, learned to be sparse, which we show in Appendix~\ref{sec:mad_visualization}. This justifies restriction of the rank promoted by the proposed representation, while direct learning of the factors reduces the memory requirements during training.

We learn the factors ${\bm\beta}^d$ and ${\bm\gamma}^d$ 
from scratch, and compute their product prior to scaling the base filter bank $\mathbf{f}$. 
The domain-specific convolution $\mathbf{g}^{d}$ is obtained as:
\begin{equation}
\label{eq:mad_new_convolution_factorised}
    [\mathbf{g}]^{d}_{mnkl} = \left(\sum_{i=1}^{I}\beta_{mi}^{d} \cdot \gamma_{in}^{d}\right) \cdot [\mathbf{f}]_{mnkl}.
\end{equation}
In this case MAD 
can still be interpreted as scaling the input channels in a specific way for each output channel, but the scaling coefficients are no longer completely independent.
This adapter factorisation reduces the number of parameters being trained from $M N$ in the full adapter to $I(M+N)$ in the factorised case. The intermediate dimension $I$ is a hyper-parameter that can be used to trade-off the number of parameters with prediction accuracy.

\subsection{Comparison to related approaches} 
\label{sec:g_comparison}
Some earlier work described in \sect{related_work} can also be understood in terms of domain specific filters $\mathbf{g}^{d}$ obtained with domain specific adapters ${\bm \alpha}^{d}$ in order to modify the filters $\mathbf{f}$ of the base network.
This allows us to further clarify the differences and similarities between these approaches and our Modulation Adapters. 
In this discussion, both $\mathbf{g}^{d}$ and $\mathbf{f}$ have the same shapes as before, the form of ${\bm \alpha}^{d}$ varies across approaches.

\emph{Masking methods} 
apply element-wise scaling of convolutional weights: 
\begin{equation}
\label{eq:masks_new_convolution}
    [\mathbf{g}]^{d}_{mnkl} = \alpha^{d}_{mnkl} \cdot [\mathbf{f}]_{mnkl},
\end{equation}
where the scaling coefficients $\alpha^d_{mnkl}$ are either binary~\cite{mallya18eccv}, or take two unique non-binary values~\cite{mancini18eccv}.
Rather than masking individual weights, BA$^2$ \cite{berriel19iccv} masks the entire features:
\begin{equation}
\label{eq:ba2_new_convolution}
    [\mathbf{g}]^{d}_{mnkl} = \alpha^{d}_{m} \cdot [\mathbf{f}]_{mnkl}.
\end{equation}
Although these adapters are compact to store, their binary nature and factored scaling of features limit task adaptation. 
Non-binary weights in our Modulation Adapters provide more degrees of freedom for adaptation, while our factorisation approach allows to control the parameter budget.

\emph{Linear feature combination methods} \cite{rosenfeld18pami} learn adapters that linearly combine filter outputs:
\begin{equation}
\label{eq:dan_new_convolution}
    [\mathbf{g}]^{d}_{mnkl} = \sum_{i=1}^{M} \alpha^{d}_{mi} \cdot [\mathbf{f}]_{inkl},
\end{equation}
which is equivalent to inserting a domain specific \myres{1} convolution after the (fixed) convolution of the base network.
Residual Adapters~\cite{rebuffi17nips}, depicted in \fig{residual_adapters}, add a skip-connection on top of  linear combination of features:
\begin{equation}
\label{eq:ra_new_convolution}
    [\mathbf{g}]^{d}_{mnkl} = \sum_{i=1}^{M} (1 + \alpha^{d}_{mi}) \cdot [\mathbf{f}]_{inkl}.
\end{equation}
Linear combination approaches have comparable parameter budget as our (non-factored) Modulation Adapters: $M^2$ for the former, and $MN$ for the latter. 
These adapters are implemented as additional linear layers applied to the output of the corresponding unaltered convolutions. Our approach can not be viewed as a linear transformation applied to either inputs or outputs, since we perform element-wise multiplication of adapters with parameters of convolutions.

\emph{Parallel Residual Adapters} \cite{rebuffi18cvpr}, depicted in \fig{parallel_adapters}, adopt a domain specific \myres{1} convolution in parallel to the fixed \myres{3} convolutions of the base network. This is equivalent to adapting the  central  element of the pre-tained filters:
\begin{equation}
\label{eq:pa_new_convolution}
    [\mathbf{g}]^{d}_{mnkl} = [\mathbf{f}]_{mnkl} + \left\{\begin{array}{ll} \alpha^{d}_{mn} & \text{if } k=l=(K-1)/2+1, \\ 0 & \text{otherwise}. \end{array} \right.
\end{equation}
While the number of adapter parameters is $MN$ as in our approach, this adapter only affects the central elements of the filters, which is restrictive and limits the space of possible adaptations.

\section{Experimental Evaluation}
\label{sec:experiments}
We describe our experimental setup in \sect{experimental_setup}. After that, we present ablation experiments in \sect{ablative_studies}, followed by comparison to previous work in \sect{sota}.

\subsection{Experimental setup} 
\label{sec:experimental_setup}

\mypar{Datasets.}
In our experiments we use the two most common multi-domain learning benchmarks: the Visual Decathlon Challenge~\cite{rebuffi17nips} and ImageNet-to-Sketch~\cite{mallya18eccv}, which contain image classification datasets from heterogeneous visual domains, and each dataset contains different classes to recognise.

The {\it Visual Decathlon Challenge} consists of ten image classification datasets: ImageNet \cite{russakovsky15ijcv}, CIFAR-100 \cite{krizhevsky09msc}, Aircraft \cite{maji13tr}, Daimler pedestrian classification \cite{munder06ieee}, Describable textures \cite{cimpoi14cvpr}, German traffic signs \cite{stallkamp12nn}, Omniglot \cite{lake15science}, Street view house numbers \cite{netzer11nips}, UCF101 Dynamic Images \cite{soomro12crcv, bilen16cvpr}, and VGG-Flowers \cite{nilsback08iccvgip}.

The second benchmark, {\it ImageNet-to-Sketch}, consists of six image classification datasets: ImageNet \cite{russakovsky15ijcv}, VGG-Flowers \cite{nilsback08iccvgip}, Stanford Cars \cite{krause13iccv}, Caltech-UCSD Birds \cite{welinder10tr}, Skteches \cite{eitz12acmtg} and WikiArt \cite{saleh15ijdah}.

For the base network architectures we follow previous work. 
For the Visual Decathlon challenge we use the ResNet26 architecture~\cite{rebuffi17nips,rosenfeld18pami,li19cvpr}, and for ImageNet-to-Sketch we use the DenseNet-121 architecture~\cite{mallya18cvpr,mancini18eccv,berriel19iccv}. 
More details on the network architectures, data processing and training details are described in Appendix~\ref{sec:network_and_training}.

\mypar{Evaluation metrics.}
We report average classification accuracy across all domains, as well as domain-specific accuracies.  Each model is trained 10 times with random initialisations, and the average across these runs is reported. 
We also use the score function introduced by Rebuffi \etal \cite{rebuffi17nips}. This score $S$ is computed as: 
\begin{equation}
    S = \sum_{d=1}^D a_{d}\;  \textrm{max} \{0, E_{d}^{\textrm{max}} - E_{d} \}^{b_{d}},
\end{equation}
where $D$ is the number of domains in the benchmark, $a_d = 1000  \left(E_{d}^{\textrm{max}}\right)^{-b_{d}}$ is the weight which ensures a perfect score of  1000 on a single domain, $E_{d}^{\textrm{max}}$ and $E_{d}$ are the baseline and the model's  test errors, respectively, and $b_{d} = 2$ is the exponent that rewards more significant reductions of the classification error.
The baseline error $E_{d}^{\textrm{max}}$ is twice the error of a per-domain fully finetuned model. 
The score of the finetuning baseline is therefore 250 per dataset. We report the total number of trainable parameters relative to the size of the base feature extractor network, not counting the linear classification head.

\begin{table*}[!th]
\centering
\resizebox{\textwidth}{!}{
\begin{tabular}{lc|cccccccccc|cc}
\toprule
 & Params     & ImNet         & Airc          & C100          & DPed          & DTD           & GTSR          & Flwr          & Oglt          & SVHN          & UCF           & Mean          & Score       \\
\midrule
Modulation adapter (full)                                                      & 2          & 60.8 &  66.8    & 79.2          & 97.9          & 56.9          & 99.2          & 86.4   & 90.0 & 97.0          & 52.5 & 78.7 $\pm$ 0.3         & 3828 $\pm$ 178   \\
Scaling central element                                                   & 2          & 60.8 & 39.6          & 74.7          & 97.7          & 53.1          & 98.0          & 76.5          & 85.8          & 94.8          & 44.0          & 72.5  $\pm$ 0.3        & 2252  $\pm$ 120         \\
Additive \myres{3} adapt.                                                        & 2          & 60.8 & 30.5          & 53.6          & 91.5          & 25.3         & 95.1          & 44.1          & 88.8          & 94.5          & 27.2          & 61.1  $\pm$ 0.2        & ~~916  $\pm$ ~~11        \\
\midrule
Modulation adapter ($I=36$)                                       
& 1.39       & 60.8          & 65.3          & 81.7          & 98.2          &  59.4          & 99.2          & 84.8          & 89.5      & 96.8          & 52.1          &  78.8 $\pm$ 0.1    & 3798 $\pm$ 134         \\
Modulation adapter (best)                                                  & 1.52       & 60.8          &  65.3          & 81.7          &  98.3    &  59.4    & 99.3          & 85.1          & 89.6    & 96.8          & {52.5} & 78.9 $\pm$ 0.1 & 3878 $\pm$ ~~68 \\
Modulation adapter (LOO)                                                   & 1.38       & 60.8          & 64.9          & 81.7          & 98.2          & 59.1          & 99.2          & 84.4          & 89.5          & 96.8          & 52.1          & 78.7 $\pm$ 0.3         & 3802 $\pm$ 123   \\
\bottomrule
\end{tabular}
}
\caption{Ablation experiments. 
Classification accuracy is reported per dataset, along with the mean, and the decathlon score.
For the latter two we also report their standard deviations across the ten repetitions of the experiments. 
}
\label{tab:ablation}
\end{table*}

\subsection{Ablation studies}
\label{sec:ablative_studies}

We conduct ablation studies on the Visual Decathlon Challenge in order to assess several variants of our model, as well as the effect of scaling, selecting the parameter budget, and the complementarity with additive adapters.

\mypar{Scaling central element  \vs entire filter.}
We experiment with a version of Modulation Adapter in which we only scale the central element of the \myres{3} filters, rather than the entire filter.
This is similar to  additive Parallel Adapters (PA) \cite{rebuffi18cvpr}, which use  \myres{1} filters to adapt the central  elements of the \myres{3} filters of the base network.
With this central-element-only scaling approach, we obtain a mean accuracy of 72.5\% and a decathlon score of 2252, see \tab{ablation}. 
This is significantly worse than the performance of our full model (78.7\% mean accuracy, and score 3828), showing the benefit of scaling the entire \myres{3} filter for each input-output channel combination.
Interestingly, scaling only the central element is also worse than the result of Parallel Adapters, which reported a mean accuracy of 78.1\% and a score of 3412 (see Table \ref{tab:vdc_comparison_table}). Although scaling the central element is equivalent after training,  our hypothesis is that optimising the scaling adapter with stochastic gradient descend is more difficult than optimising the additive adapter. 
Using an optimiser with adaptive learning rates, such as Adam~\cite{kingma15iclr}, may alleviate this. 

\mypar{Additive adaptation of all filter elements.}
In our second ablation, we consider a variant of our approach in which we adapt all filter elements, as in \Eq{mad_new_convolution}, but in an additive rather than multiplicative manner. 
This is similar to PA, but now the adapter updates all the filter elements rather than only the central one. 
This variant leads to a substantially worse mean accuracy of 61.1\% and a decathlon score of 916, see line ``Additive \myres{3} adapt.'' in \tab{ablation}.
Our interpretation of this deterioration is that strong additive adaptation of all filter weights leads to spatially uniform filters, that fail to detect spatial patterns. 

\begin{figure}[t]
\begin{center}
\includegraphics[width=0.9\columnwidth]{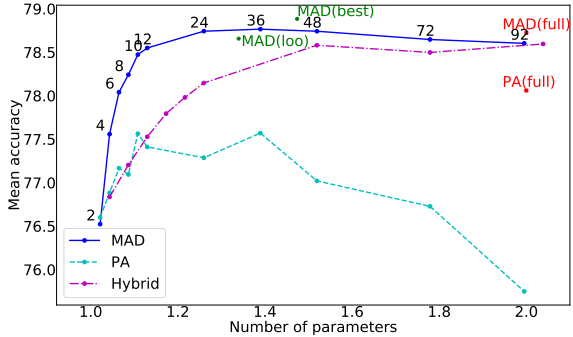}
\end{center}
\caption{
Mean accuracy \vs  parameter budget for different adapter-based methods:   decomposed version of our Modulation Adapters (MAD), decomposed version of Parallel Adapters (PA, our implementation), hybrid adapters that combine MAD and PA.
}
\label{fig:mad_ablations_acc}
\end{figure}

\mypar{Decomposition of Modulation Adapters.}
In \fig{mad_ablations_acc}, we consider the effect of decomposing our Modulation Adapters as a product of two smaller matrices of weights, as in \Eq{mad_decomposition}.
We vary the intermediate dimension $I$ from 2 to 92, and plot the mean accuracy against the parameter budget. 
We note that the mean accuracy quickly increases when allowing more parameters, reaches a maximum at $I=36$, and then slightly declines.
The mean accuracy of the full, non-decomposed model, and the model with $I=36$ are very close at 78.7\% and 78.8\% respectively, see \tab{ablation}. 
The decomposed model with $I=36$, however, yields a saving of more than 60\% in the parameter budget.

While on average there is a smooth trend between the accuracy and the parameter budget, the optimal parameter budget  differs per dataset.
In \tab{ablation} (line ``Modulation Adapter (best)''), we show what the optimal accuracy would be if we were to set this budget for each dataset separately. 
We notice that this only leads to minor improvements of 0.1\% and 0.2\% \wrt the $I=36$ and the full models, which is within the standard deviation of the reported results. 
We further illustrate the stability of the trade-off between accuracy and parameter budget
across datasets with a leave-one-out (LOO) experiment. In this case, for each dataset we select the parameter budget by taking the optimal budget on all the other datasets. We then average the results across all the datasets. 
As in previous cases, we only find minor deviations in the mean accuracy, reaching 78.7\%, which is the same as the performance of the full non-decomposed model.

\mypar{Decomposition of Parallel Adapters.}
Similar to the decomposition used in our Modulation Adapters, we implemented a decomposition version of Parallel Adapters (PA)~\cite{rebuffi18cvpr}, where each \myres{1} adapter is given as a product of two matrices with smaller intermediate dimension $I$. 
As shown in \fig{mad_ablations_acc}, unlike our approach, PA does not behave well with  decomposition relative to the full version of PA.
Moreover, decomposed PA performs consistently worse than our Modulation Adapters across all parameter budgets.
For reference, we also included  the ``full'' non-decomposed version of MAD and PA in the graph.\footnote{ Result for PA (full) taken from original paper: $78.1\%$ mean accuracy and score $3412$.
We implemented  the decomposed version of PA  (as well as our models) based on the public code-base of PA (full) from \cite{rebuffi18cvpr}. In our experiments, 
we obtained mean accuracy $77.7\%$ and score $3206$ for PA (full). 
If we take best results across these ten runs, we obtain mean accuracy $78.2\%$ and score $3460$, consistent with  the original paper.
}

\mypar{Hybrid adapters.}
To study the complementarity of multiplicative and additive adaptation, we experiment with a hybrid approach 
that mixes our Modulation Adapters with Parallel Adapters \cite{rebuffi18cvpr}.
We use decomposed versions of MAD and PA, since our goal is to keep the total number of parameters below the size of the full version of a single model. 
The available parameter budget is evenly split between the two types of adapters. 
The results in \fig{mad_ablations_acc} show that these adapters are not complementary:
the hybrid approach is consistently worse than our  Modulation Adapters, in particular for small parameter budgets, while  providing a consistent improvement over the decomposed version of Parallel Adapters.

\begin{figure}
\centering
\includegraphics[width=0.92\columnwidth]{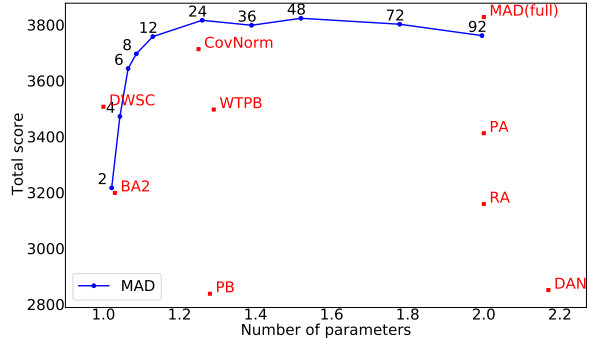}
\caption{
Total score \vs parameter budget for Modulation Adapters (MAD) and the other approaches. 
}
\label{fig:vdc_comparison_score}
\end{figure}

\begin{table*}[!t]
\centering
\begin{small}
\resizebox{\textwidth}{!}{
\begin{tabular}{lc|cccccccccc|ccc}
\toprule
 & Params     & ImNet         & Airc          & C100          & DPed          & DTD           & GTSR          & Flwr          & Oglt          & SVHN          & UCF           & Mean          & Score       & Rank       \\
Number of images & & 1.3m & 7k & 50k & 30k & 4k & 40k & 2k & 26k & 70k & 9k\\
\midrule
Feature \cite{rebuffi17nips}                                                       & 1 & 59.7          & 23.3          & 63.1          & 80.3          & 45.4          & 68.2          & 73.7          & 58.8          & 43.5          & 26.8          & 54.3          & 544           & 14.7          \\
BN Adapt.\ \cite{bilen17arxiv} & $\sim$1 & 59.9          & 43.1           & 78.6           & 92.1       & 51.6          & 95.8          & 74.1          & 84.8          & 94.1          & 43.5          & 71.8          & 1363          & 13.4          \\
BA2 \cite{berriel19iccv}                                                    & 1.03 & 56.9          & 49.9          & 78.1          & 95.5          & 55.1          & {\ul 99.4}          & 86.1          & 88.7          & 96.9          & 50.2          & 75.7          & 3199          & 8.9          \\
Finetune \cite{rebuffi17nips}                                                       & 10         & 59.9          & 60.3          & {\bf 82.1}          & 92.8          & 55.5          & 97.5          & 81.4          & 87.7          & 96.6          & 51.2          & 76.5          & 2500         & 9.4          \\
PB \cite{mallya18eccv}                                                       & 1.28       & 57.7          & 65.3          & 79.9          & 97.0          & 57.5          & 97.3          & 79.1          & 87.6          & \textbf{97.2} & 47.5          & 76.6          & 2838          & 9.1          \\
DAN \cite{rosenfeld18pami}                                                   & 2.17       & 57.7          & 64.1          & 80.1          & 91.3          & 56.5          & 98.5          & 86.1          & {\ul 89.7} & 96.8          & 49.4          & 77.0          & 2851          & 8.3          \\
RA       \cite{rebuffi17nips}                                                      & 2          & 60.3          & 61.9          &  81.2    & 93.9          & 57.1          & 99.3          & 81.7          & 89.6          & 96.6          & 50.1          & 77.2          & 3159          & 7.6          \\
WTPB (full) \cite{mancini18eccv}                                                  & 1.29       & {\ul 60.8} & 52.8          & {\ul 82.0} & 96.2          & 58.7          & 99.2          & {\bf 88.2} & 89.2          & 96.8          & 48.6          & 77.2          & 3497          & 5.8          \\
DWSC \cite{guo19aaai}                                                     & $\sim$1 & {\bf 64.0}          & 61.1          & 81.2          & 97.0          & 55.5          & 99.3          & 85.7          & 89.1          & 96.2          & 49.3          & 77.8          & 3507          & 7.4          \\
SpotTune \cite{guo19iccv}                                                       & 11         & 60.3          & 63.9          & 80.5          & 96.5          & 57.1          & {\bf 99.5} & 85.2          & 88.8          & 96.7          & {\ul 52.3}    & 78.1          & 3612          & 6.4          \\
PA \cite{rebuffi18cvpr}                                                      & 2          & 60.3          & 64.2          &  81.9 & 94.7          & 58.8          &  {\ul 99.4}          & 84.7          & 89.2          & 96.5          & 50.9          & 78.1          & 3412          & 6.5          \\
CovNorm \cite{li19cvpr}                                                      & 1.25       & 60.4       & {\bf 69.4} & 81.3          & {\bf 98.8} & {\bf 59.9} & 99.1          & 83.4          & 87.7          & 96.6          & 48.9          & 78.6          & 3713          & 6.6          \\
\midrule
MAD (full)                                                      & 2          & {\ul 60.8} & {\ul 66.8}    & 79.2          & 97.9          & 56.9          & 99.2          & {\ul 86.4}    & {\bf 90.0} & {\ul 97.0}    & {\bf 52.5} & {\ul 78.7}          & {\bf 3828}        & 4.1           \\
MAD ($I=24$)                                        
& 1.26       & {\ul 60.8}          & 64.9          &  81.5          & {\ul 98.2}          & {\ul 59.1}          & 99.2          & 85.1          & 89.5        & 96.8          & {\ul 52.3}          & {\ul 78.7}    & {\ul 3816}          & {\ul 4}          \\
MAD ($I=36$)                                        
& 1.39       & {\ul 60.8}          & 65.3          &  81.7          & {\ul 98.2}          & {\ul 59.4}          & 99.2          & 84.8          & 89.5        & 96.8          & 52.1          & {\bf 78.8}    & 3798          & {\bf 3.9}          \\   
\bottomrule
\end{tabular}
}
\end{small}
\caption{
Comparison to the state-of-the-art methods on the Visual Decathlon Challenge.
Best results per metric 
(mean and per dataset classification accuracy, as well as score) are  in bold, and the second best are underlined.
} 
\label{tab:vdc_comparison_table}
\end{table*}

\subsection{Comparison to the state of the art}
\label{sec:sota}

\mypar{Visual Decathlon Challenge.}
In \fig{vdc_comparison_score}, we compare the total score of our full and decomposed Modulation Adapters to state-of-the-art methods across a range of parameter budgets. 
While other methods provide a single operating point, our decomposed Modulation Adapters enable adaptation to new domains across a wide range of parameter budgets, and yield better or comparable score.
In particular, CovNorm~\cite{li19cvpr} has a score $3713$,
and for a similar parameter budget our Modulation Adapters ($I\!=\!24$)  obtain an improved score $3816$.
Parallel Adapters~\cite{rebuffi18cvpr} report a score $3412$, while at the same parameter budget our non-decomposed Modulation Adapters obtain a score $3828$.
With $3507$, DWSC~\cite{guo19aaai} achieves a lower score than the top-performing methods, but uses a smaller parameter budget. Unlike the other methods, however, DWSC does not use the ResNet-26 base network, but an architecture based on depthwise separable convolutions, and is therefore not directly comparable. The number of parameters is still given relative to the ResNet-26 model.

In \tab{vdc_comparison_table}, we additionally report the accuracy per dataset as well as the decathlon score. Here, we also include three more baselines.
The ``Feature'' baseline does not adapt the base network, and only learns the linear classification head per dataset. This baseline obtains poor mean accuracy (54.3\%) and score (544), due to insufficient adaptation.
The ``BN Adapt.'' baseline only learns dataset-specific BatchNorm layers, but leaves all the other parameters of the base network unchanged.
This adds very few parameters, but leads to a remarkable improvement with respect to the feature baseline, by  scaling the output of the convolutional layers in a dataset-specific manner. 
After training, this approach is equivalent to only scaling the output channels of the filters.
Our approach is similar, but scales filter weights in both input and output dependent manner, leading to more adaptation and far superior results.
The ``Finetune'' baseline finetunes all the parameters of the base network for each dataset, and  does not offer any savings in the parameter budget. It is outperformed by other methods that limit the number of free parameters,  allowing for better generalisation to datasets with few samples.
We also include SpotTune, which improves over the Finetune baseline in prediction accuracy, but does not offer reduction in the number of parameters.

Our Modulation Adapters (MAD) not only obtain high mean accuracy and score, but also offer the best or comparable accuracy per domain.
This is unlike the other methods whose performance is often less consistent across datasets.
To quantify this, we rank the methods by accuracy (best first)  on each  domain, and then average the rank of each method across domains.
On the Visual Decathlon Challenge, MAD (I=36), MAD (I=24) and MAD (full) have average ranks 3.9, 4.0 and 4.1, respectively. These are the best three results on this benchmark. 
CovNorm, which is our next competitor in terms of the score and accuracy, has the average rank 6.6, which is the seventh-best result. 
Furthermore, MAD is significantly easier to train than CovNorm. The latter consists of a complex multi-step algorithm: (i) learning Residual Adapters~\cite{rebuffi17nips}, (ii) computing PCA for input and output activations, (iii) learning additional mini-adaptation layers, and (iv) joint finetuning of all components of the final adapters. 
Training our model is comparable to the first step of CovNorm: 
we learn a single multiplicative adapter per domain in a single training run.

\begin{table*}[t]
\centering
\begin{small}
\resizebox{\textwidth}{!}{
\begin{tabular}{lc|cccccc|ccc}
\toprule
 & Params     & ImNet         & CUBS         & Stanford Cars          & Flowers          & WikiArt           & Sketch          & Mean          & Score        & Rank           \\
Number of images & & 1.3m & 6k & 8k & 2k & 42k & 16k \\
\midrule
Feature \cite{mallya18eccv}                                                        & 1 & {\bf 74.4}          & 73.5          & 56.8          & 83.4          & 54.9          & 53.1          & 66.0          & 324           & 6.8          \\
PB \cite{mallya18eccv}                                                       & 1.21       & {\bf 74.4}          & 81.4          & 90.1         & 95.5          & 73.9          & 79.1          & 82.4          & 1209          & 5.8          \\
BA2 \cite{berriel19iccv}                                                    & 1.17 & {\bf 74.4}          & 82.4           & {\bf 92.9}          & 96.0          & 71.5          & 79.9          & 82.9          & 1434          & 4          \\
WTPB (full) \cite{mancini18eccv}                                                   & 1.21       & {\bf 74.4}        & 81.7          & 91.6            & {\bf 96.9}          & 75.7          & 79.8          & 83.4          & 1534          & 3.5          \\
Finetune \cite{mallya18eccv}                                                        & 6         & {\bf 74.4}          & 81.9          & 91.4          & {\ul 96.5}          & {\ul 76.4}          & {\ul 80.5}          &  83.5         & 1500         & 3          \\
\midrule
MAD (full)                                                      & 4.61          & {\bf 74.4}          & {\bf 83.9}          & {\ul 91.9}          & {\bf 96.9}          & {\bf 76.9}          & {\bf 81.0}          & {\bf 84.2}          & {\bf 1668}           &  {\bf 1.2}          \\
MAD ($I$=36)                                                      & 2.33          & {\bf 74.4}          & {\ul 83.1}              &  90.6              & 96.4          & 75.3              & 80.2            &  83.3             & 1446              &  3.7             \\
MAD (full) + WTPB (full)                                      
& 1.26       & {\bf 74.4}          & 82.7          &  91.4          & {\bf 96.9}          & 76.2          & 80.1          & {\ul 83.6}            & {\ul 1569}          & {\ul 2.7}          \\
\bottomrule
\end{tabular}
}
\end{small}
\caption{
Comparison to the state-of-the-art methods on the ImageNet-to-Sketch benchmark.
Best results per metric 
(mean and per dataset classification accuracy, as well as score) are  in bold, and the second best are underlined.
}
\label{tab:its_comparison_table}
\end{table*}

\mypar{ImageNet-to-Sketch benchmark.}
In \tab{its_comparison_table}, we compare our Modulation Adapters with the state of the art on the ImageNet-to-Sketch benchmark.
The full (non-decomposed) model outperforms all the other methods both in terms of the average accuracy and the total score, and is the only one to outperform finetuning the base network (``Finetune"). 
This result further confirms the effectiveness of the proposed multiplicative adaptation strategy.
While our parameter budget is larger than those of our competitors (due to abundance of \myres{1} convolutions in the DenseNet-121 base network), it is still below the budget of the fully finetuned models, which allows us to reduce the performance gap between the latter and learning only domain-specific classification heads (``Feature"). As for individual tasks, Modulation Adapters demonstrate the best performance on four datasets (CUBS, Flowers, WikiArt and Sketch), and the second-best on Stanford Cars. 
Results on WikiArt and Sketch are of special interest, since these datasets are quite  different from ImageNet which was used for pre-training. While the latter consists of natural images, the former two contain paintings and sketches, which makes the multi-domain task on them more challenging. 
On ImageNet-to-Sketch, MAD (full) obtains the best average rank 1.2, while next competitors ``Finetune''and  WTPB obtain 3 and 3.5.

We found that factorisation of our Modulation Adapters is less effective on the ImageNet-to-Sketch dataset: using $I=36$ intermediate dimensions for decomposition (reducing the parameter budget roughly by a factor two), we obtained an average accuracy of $83.3$.
The different behaviour of decomposition across the two benchmarks may be related to the relatively large number of parameters in \myres{1} convolutions in the DenseNet-121 architecture, whereas in the ResNet-26 architecture most parameters reside in \myres{3} convolutions. We also found that the learned adapters for the ImageNet-to-Sketch benchmark do not show the sparse structure that we observe in the adapters learned for the Visual Decathlon Challenge, see Appendix~\ref{sec:mad_visualization}, making rank restriction less effective. Despite the less consistent performance, MAD ($I$=36) still outperforms other competitors on Sketch and CUBS, improving over full finetuning of the base network on the latter.

As an alternative strategy to reduce the parameter budget, we explored the  possibility of combining our method with other approaches. 
In particular, we trained a hybrid model where WTPB~\cite{mancini18eccv} is applied to \myres{1} convolutions, while Modulation Adapters are applied to \myres{3} convolutions (``MAD (full) + WTPB (full)''). The hybrid model improves over WTPB on three datasets, bringing it closer to individual finetuning on WikiArt and Sketch, and even outperforming it on CUBS. 
In terms of the average accuracy, the total score and the average rank, the hybrid model is slightly better than individual finetuning, second only to our Modulation Adapters. This shows that even more flexible models could be created from existing approaches depending on the desired parameter budget and performance requirements.

\section{Conclusion}
In this paper, we address the problem of learning across multiple domains while reducing the total parameter budget. 
In contrast to previous works which rely on adaptation modules that are limited in their capacity, or only adapt a part of the base network parameters, we introduce Modulation Adapters, a novel type of adapters based on flexible weight modulation. 
We design these adapters to update the weights of the pre-trained convolutional filters by scaling their input and output channels, and provide an efficient parameterisation through decomposition for further parameter reduction. 
Ablative analyses as well as our model's excellent performance on the popular Visual Decathlon Challenge and ImageNet-to-Sketch benchmarks validate the benefits of our multiplicative adaptation method.
At the same time other means of parameter reduction for the full Modulation Adapter could be considered to cover the setups where the learned adapters are not sparse.

{\small
\bibliographystyle{ieee_fullname}
\bibliography{main}

\begin{thebibliography}{10}\itemsep=-1pt

\bibitem{anokhin21cvpr}
I. Anokhin, K. Demochkin, T. Khakhulin, G. Sterkin, V. Lempitsky, and D.
  Korzhenkov.
\newblock Image generators with conditionally-independent pixel synthesis.
\newblock In {\em CVPR}, 2021.

\bibitem{noguchi19iccv}
N. Atsuhiro and H. Tatsuya.
\newblock Image generation from small datasets via batch statistics adaptation.
\newblock In {\em ICCV}, 2019.

\bibitem{berriel19iccv}
R. Berriel, S. Lathuillere, M. Nabi, T. Klein, T. Oliveira-Santos, N. Sebe, and
  E. Ricci.
\newblock Budget-aware adapters for multi-domain learning.
\newblock In {\em ICCV}, 2019.

\bibitem{bilen16cvpr}
H. Bilen and A. Vedaldi.
\newblock Weakly supervised deep detection networks.
\newblock In {\em CVPR}, 2016.

\bibitem{bilen17arxiv}
H. Bilen and A. Vedaldi.
\newblock Universal representations: The missing link between faces, text,
  planktons, and cat breeds.
\newblock {\em arXiv preprint}, 2017.

\bibitem{chollet17cvpr}
F. Chollet.
\newblock Xception: Deep learning with depthwise separable convolutions.
\newblock In {\em CVPR}, 2017.

\bibitem{cimpoi14cvpr}
M. Cimpoi, S. Maji, I. Kokkinos, S. Mohamed, and A. Vedaldi.
\newblock Describing textures in the wild.
\newblock In {\em CVPR}, 2014.

\bibitem{eitz12acmtg}
M. Eitz, J. Hays, and M. Alexa.
\newblock How do humans sketch objects?
\newblock {\em ACM Trans.\ Graphics}, 31(4):1--10, 2012.

\bibitem{robb20arxiv}
R. Esther, C. Wen-Sheng, K. Abhishek, and H. Jia-Bi.
\newblock Few-shot adaptation of generative adversarial networks.
\newblock {\em arXiv preprint}, 2020.

\bibitem{guo19aaai}
Y. Guo, Y. Li, L. Wang, and T. Rosing.
\newblock Depthwise convolution is all you need for learning multiple visual
  domains.
\newblock In {\em AAAI}, 2019.

\bibitem{guo19iccv}
Y. Guo, H. Shi, A. Kumar, K. Grauman, T. Rosing, and R. Feris.
\newblock {SpotTune}: {T}ransfer learning through adaptive fine-tuning.
\newblock In {\em ICCV}, 2019.

\bibitem{he16cvpr}
K. He, X. Zhang, S. Ren, and J. Sun.
\newblock Deep residual learning for image recognition.
\newblock In {\em CVPR}, 2016.

\bibitem{he16eccv}
K. He, X. Zhang, S. Ren, and J. Sun.
\newblock Identity mappings in deep residual networks.
\newblock In {\em ECCV}, 2016.

\bibitem{houlsby19icml}
N. Houlsby, A. Giurgiu, S. Jastrzebski, B. Morrone, Q.~De Laroussilhe, A.
  Gesmundo, M. Attariyan, and S. Gelly.
\newblock Parameter-efficient transfer learning for {NLP}.
\newblock In {\em ICML}, 2019.

\bibitem{karras20cvpr}
T. Karras, S. Laine, M. Aittala, J. Hellsten, J. Lehtinen, and T. Aila.
\newblock Analyzing and improving the image quality of stylegan.
\newblock In {\em CVPR}, 2020.

\bibitem{kingma15iclr}
D. Kingma and J. Ba.
\newblock Adam: A method for stochastic optimization.
\newblock In {\em ICLR}, 2015.

\bibitem{krause13iccv}
J. Krause, M. Stark, J. Deng, and L. Fei-Fei.
\newblock 3d object representations for fine-grained categorization.
\newblock In {\em ICCV Workshops}, 2013.

\bibitem{krizhevsky09msc}
A. Krizhevsky.
\newblock Learning multiple layers of features from tiny images.
\newblock Master's thesis, University of Toronto, 2009.

\bibitem{lake15science}
B.~M. Lake, R. Salakhutdinov, and J.~B. Tenenbaum.
\newblock Human-level concept learning through probabilistic program induction.
\newblock {\em Science}, 350(6266):1332--1338, 2015.

\bibitem{lecun15nature}
Y. LeCun, Y. Bengio, and G. Hinton.
\newblock Deep learning.
\newblock {\em Nature}, 52:436--444, 2015.

\bibitem{li19cvpr}
Y. Li and N. Vasconcelos.
\newblock Efficient multi-domain learning by covariance normalization.
\newblock In {\em CVPR}, 2019.

\bibitem{loshchilov19iclr}
I. Loshchilov and F. Hutter.
\newblock Decoupled weight decay regularization.
\newblock In {\em ICLR}, 2019.

\bibitem{maji13tr}
S. Maji, J. Kannala, E. Rahtu, M. Blaschko, and A. Vedaldi.
\newblock Fine-grained visual classification of aircraft.
\newblock {\em arXiv preprint}, arXiv:1306.5151, 2013.

\bibitem{mallya18eccv}
A. Mallya, D. Davis, and S. Lazebnik.
\newblock Piggyback: Adapting a single network to multiple tasks by learning to
  mask weights.
\newblock In {\em ECCV}, 2018.

\bibitem{mallya18cvpr}
A. Mallya and S. Lazebnik.
\newblock Packnet: Adding multiple tasks to a single network by iterative
  pruning.
\newblock In {\em CVPR}, 2018.

\bibitem{mancini18eccv}
M. Mancini, E. Ricci, B. Caputo, and S.~Rota Bulo.
\newblock Adding new tasks to a single network with weight transformations
  using binary masks.
\newblock In {\em ECCV Workshops}, 2018.

\bibitem{munder06ieee}
S. Munder and D.~M. Gavrila.
\newblock An experimental study on pedestrian classification.
\newblock {\em PAMI}, 28(11):1863--1868, 2006.

\bibitem{netzer11nips}
Y. Netzer, T. Wang, A. Coates, A. Bissacco, B. Wu, and A.~Y. Ng.
\newblock Reading digits in natural images with unsupervised feature learning.
\newblock In {\em NeurIPS Workshop on Deep Learning and Unsupervised Feature
  Learning}, 2011.

\bibitem{nilsback08iccvgip}
M.-E. Nilsback and A. Zisserman.
\newblock Automated flower classification over a large number of classes.
\newblock In {\em ICVGIP}, 2008.

\bibitem{rebuffi17nips}
S.-A. Rebuffi, H. Bilen, and A. Vedaldi.
\newblock Learning multiple visual domains with residual adapters.
\newblock In {\em NeurIPS}, 2017.

\bibitem{rebuffi18cvpr}
S.-A. Rebuffi, H. Bilen, and A. Vedaldi.
\newblock Efficient parametrization of multi-domain deep neural networks.
\newblock In {\em CVPR}, 2018.

\bibitem{rosenfeld18pami}
A. Rosenfeld and J.K. Tsotsos.
\newblock Incremental learning through deep adaptation.
\newblock {\em PAMI}, 42(3):651--663, 2018.

\bibitem{russakovsky15ijcv}
O. Russakovsky, J. Deng, H. Su, J. Krause, S. Satheesh, S. Ma, Z. Huang, A.
  Karpathy, A. Khosla, and M. Bernstein.
\newblock Imagenet large scale visual recognition challenge.
\newblock {\em IJCV}, 115(3):211--252, 2015.

\bibitem{saleh15ijdah}
B. Saleh and A. Elgammal.
\newblock Large-scale classification of fine-art paintings: Learning the right
  metric on the right feature.
\newblock {\em International Journal for Digital Art History}, 2016.

\bibitem{senhaji21icpr}
A. Senhaji, J. Raitoharju, M. Gabbouj, and A. Iosifidis.
\newblock Not all domains are equally complex: Adaptive multi-domain learning.
\newblock In {\em ICPR}, 2021.

\bibitem{soomro12crcv}
K. Soomro, A.~R. Zamir, and M. Shah.
\newblock A dataset of 101 human action classes from videos in the wild.
\newblock {\em Center for Research in Computer Vision}, 2(11), 2012.

\bibitem{stallkamp12nn}
J. Stallkamp, M. Schlipsing, J. Salmen, and C. Igel.
\newblock Man vs. computer: Benchmarking machine learning algorithms for
  traffic sign recognition.
\newblock {\em Neural Networks}, 32:323--332, 2012.

\bibitem{vaswani17nips}
A. Vaswani, N. Shazeer, N. Parmar, J. Uszkoreit, L. Jones, A. Gomez, L. Kaiser,
  and I. Polosukhin.
\newblock Attention is all you need.
\newblock In {\em NeurIPS}, 2017.

\bibitem{welinder10tr}
P. Welinder, S.Branson, T. Mita, C. Wah, F. Schroff, S. Belongie, and P.
  Perona.
\newblock Caltech-ucsd birds 200.
\newblock Technical report, California Institute of Technology, 2010.

\end{thebibliography}
}

\clearpage

\appendix

\begin{figure*}[!ht]
\centering
\begin{subfigure}[t]{\textwidth}
    \centering
    \includegraphics[width=\textwidth]{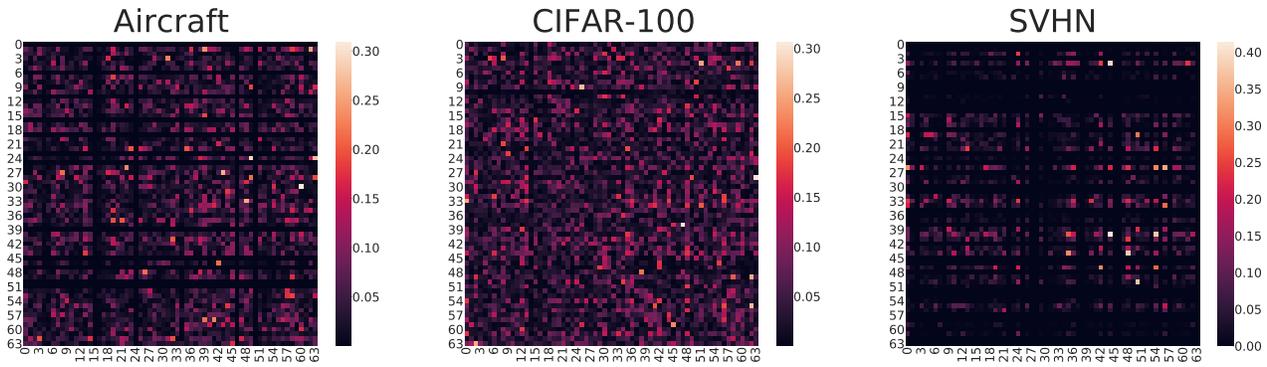}
    \caption{ResNet-26 modulation adapters learned on Visual Decathlon Challenge datasets. } 
    \label{fig:mad_weight_vis_decathlon}
\end{subfigure}
\vspace{30pt}
\begin{subfigure}[t]{\textwidth}
    \centering
    \includegraphics[width=\textwidth]{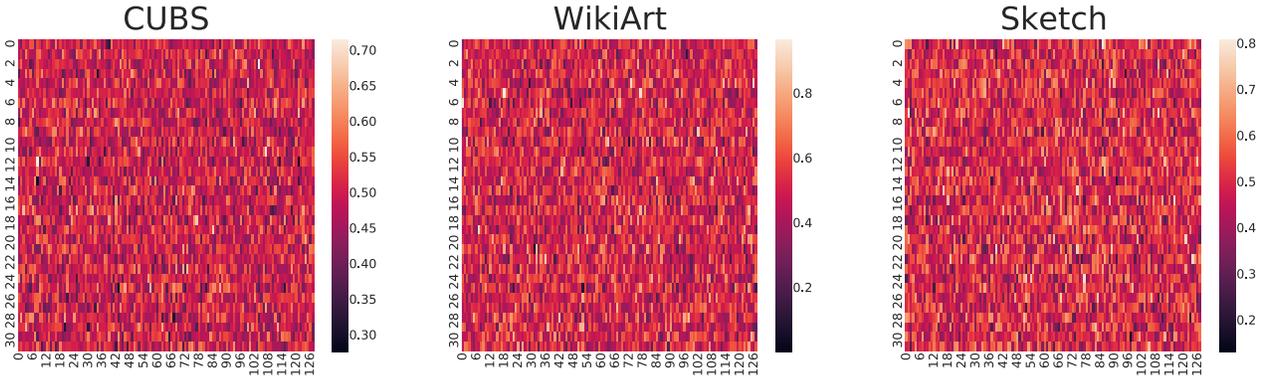}
    \caption{DenseNet-121 modulation adapters learned on ImageNet-to-Sketch benchmark datasets. }
    \label{fig:mad_weight_vis_sketch}
\end{subfigure}
\caption{Visualization of the absolute weight values of Modulation Adapters. The vertical axis represents the output channels, while the horizontal one represents the input channels.
}
\label{fig:mad_weight_vis}
\end{figure*}

In this supplementary material, we describe the experimental setup in Appendix \ref{sec:network_and_training}. 
In Appendix \ref{sec:mad_visualization}, we provide visualization of our Modulation Adapters for some of the convolutions layers in the base network on the two benchmarks.

\section{Network architecture and training}
\label{sec:network_and_training}

\paragraph{Visual Decathlon Challenge.} For the base network, we use ResNet-26~\cite{rebuffi17nips,rosenfeld18pami,li19cvpr} and the weights of its feature extractor pre-trained on ImageNet from \cite{rebuffi17nips}. All convolutional layers in the feature extractor consist of \myres{3} filters, which are fixed after pre-training and accompanied by domain-specific Modulation Adapters. The latter are initialized with ones and learned for each domain separately. Batch normalization layers, which follow each of these modulated convolutions, are also finetuned for each domain separately, and are initialized with values from the pre-trained base network. 
Finally, domain-specific fully connected layers used as classification heads are learned from scratch, individually for each task.
The model is trained with SGD with momentum 0.9 for 140 epochs. Initial learning rate is set to 0.1, and decreased by a factor ten after epochs 80 and 110. We set the weight decay to 0.0035 for VGG-Flowers, 0.0025 for Aircraft and Describable textures, 0.0015 for UCF101 Dynamic Images, 0.001 for Daimler pedestrian classification and Omniglot, and 0.0005 for CIFAR-100, German traffic signs and Street View House Numbers.
All images are resized isotropically to have a shorter size of 72 pixels. We use official data splits from \cite{rebuffi17nips}, as well as their data processing for all datasets except for Daimler pedestrian classification. For the latter, we use data processing from \cite{guo19iccv}.

\paragraph{ImageNet-to-Sketch.} For this benchmark, we use DenseNet-121 \cite{he16cvpr} as the base network, which is commonly used in previous work \cite{mallya18eccv,mancini18eccv,berriel19iccv,guo19iccv}. Similarly to the Visual Decathlon Challenge, we apply Modulation Adapters to all convolutions in the base network, including those with \myres{1} filters. Despite the absence of parameter reduction in the latter case, we empirically find that modulation of such convolutions leads to better performance compared to finetuning them directly. 
Domain-specific batch normalization layers and classification heads are initialized and trained the same way as for ResNet-26, while Modulation Adapters $\alpha$ are initialized with 0.15. The model is trained with AdamW \cite{loshchilov19iclr} for 60 epochs. 
The initial learning rate is set to 0.001, and decreased by a factor ten after epochs 20 and 40. We set the weight decay to 0.0035 for Flowers, 0.001 for CUBS, Stanford Cars and Sketch, and 0.0005 for WikiArt.
We use original data splits and data processing from \cite{mallya18eccv}. Preprocessed images have resolution \myres{224}.\\

\section{Visualization of Modulation Adapters}
\label{sec:mad_visualization}

\fig{mad_weight_vis}(a) shows the absolute weight values of Modulation Adapters (full) which adapt the same \myres{3} convolutional layer at depth five in ResNet-26 on Aircraft, CIFAR-100 and SVHN datasets from the Visual Decathlon Challenge. \fig{mad_weight_vis}(b) does the same for \myres{3} convolutional layer at depth nine in DenseNet-121 on CUBS, WikiArt and Sketch datasets from the ImageNet-to-Sketch benchmark.

Non-trivial adapters are learned, \ie the matrices do not contain  rows or columns with uniform values. In addition to that, we can observe some feature selection happening in ResNet-26: dark spots indicate zeroing of corresponding input feature maps.  
This is not the case for DenseNet-121 used for ImageNet-to-Sketch benchmark. The low sparsity of Modulation Adapters on the DenseNet-121 might explain why the decomposed version is less effective on this benchmark.

\end{document}